# Video Transformer for Deepfake Detection with Incremental Learning

Sohail Ahmed Khan
sohail.khan@mbzuai.ac.ae
Mohamed bin Zayed University of Artificial Intelligence
Abu Dhabi, United Arab Emirates

Hang Dai*
hang.dai@mbzuai.ac.ae
Mohamed bin Zayed University of Artificial Intelligence
Abu Dhabi, United Arab Emirates

## ABSTRACT

Face forgery by deepfake is widely spread over the internet and this raises severe societal concerns. In this paper, we propose a novel video transformer with incremental learning for detecting deepfake videos. To better align the input face images, we use a 3D face reconstruction method to generate UV texture from a single input face image. The aligned face image can also provide pose, eyes blink and mouth movement information that cannot be perceived in the UV texture image, so we use both face images and their UV texture maps to extract the image features. We present an incremental learning strategy to fine-tune the proposed model on a smaller amount of data and achieve better deepfake detection performance. The comprehensive experiments on various public deepfake datasets demonstrate that the proposed video transformer model with incremental learning achieves state-of-the-art performance in the deepfake video detection task with enhanced feature learning from the sequenced data.

## CCS CONCEPTS

• **Security and privacy** → **Social aspects of security and privacy**.

## KEYWORDS

Deepfakes detection, face forensics, transformer, video analysis

## 1 INTRODUCTION

Recent developments in deep learning and the availability of large scale datasets have led to powerful deep generative models that can generate highly realistic synthetic videos. State-of-the-art generative models have enormous amount of advantageous applications, but the generative models are also used for malicious purposes. One such application of the generative models is deepfake video generation. Generative models have evolved to an extent that, it is difficult to classify the real and the fake videos. Deepfake can be used for unethical and malicious purposes, for example, spreading false propaganda, impersonating political leaders saying or doing unethical things, and defaming innocent individuals. Deepfake can be grouped into four categories: face replacement, facial re-enactment, face editing, and complete face synthesis [37].

Deepfake generation techniques are increasing exponentially and becoming more and more difficult to detect. Current detection systems are not in a capacity to detect manipulated media effectively. In Deepfake Detection Challenge (DFDC) [21], the models achieve much worse performance when tested on unseen data than that on the DFDC test set. Generalization capability is one of the major concerns in the existing deepfake detection systems [10, 54]. A wide variety of detection systems [1, 3, 13, 14, 27, 30, 54] employ CNNs and recurrent networks to detect manipulated media. Li *et al.* [31] employ CNNs to detect face warping artifacts in images from the deepfake datasets [29, 55]. The proposed approach works well in cases where there are visible face warping artifacts [46]. Most of the deepfake generation techniques employ post-processing procedures to remove the warping artifacts [51], which makes it more difficult to detect deepfake videos. Another limitation of the existing approaches is that, most of the proposed systems make predictions on the frames in a video and average the predictions in order to get a final prediction score for the whole video. So it fails to consider the relationships among frames. To overcome this, we propose a novel video transformer to extract spatial features with the temporal information [19, 27, 47]. Transformers were first proposed for natural language processing tasks, by Vaswani *et al.*, in [50]. Since then, transformers show powerful performance in the natural language processing tasks, for example, machine translation, text classification [41], question-answering, and natural language understanding. The widely used transformer architectures include Bidirectional Encoder Representations from Transformers (BERT) [20], Robustly Optimized BERT Pre-training (RoBERTa) [33], Generative Pre-trained Transformer (GPT) v1-v3 [7, 44, 45]. The transformer models can naturally accommodate the video sequences for the feature learning [28].

To extract more informative features, we train our models on the aligned facial images and their corresponding UV texture maps [16, 17, 26]. The existing methods use aligned 2D face images. Such an alignment only centralizes the face without considering whether the face is frontalized. When the face is not frontalized, the face part that is not captured by the camera can cause facial information loss and misalignment with the face images that are frontalized. With the UV texture, all face images are aligned into the UV map that is created from the generated 3D faces. Since the generated 3D faces cover all the facial parts, there is no information loss. In UV map, the facial part for all the faces can be located in the same spatial space. For example, all the nose parts are located in the same region on the UV map. So the faces in UV maps are better aligned. To deal with the input combination of face image and UV texture map, we use learnable segment embeddings in the input data structure. The segment embeddings help the model to distinguish different types of inputs in the same data structure. Furthermore, we use an incremental learning strategy for finetuning our models on different datasets incrementally to achieve state-of-the-art performance on new datasets while maintaining the performance on the previous datasets.

Our contributions can be summarized in three-fold:

- We propose a video transformer with face UV texture map for deepfake detection. The experimental results on five different public datasets show that our method achieves better performance than state-of-the-art methods.
- The proposed segment embedding enables the network to extract more informative features, thereby improving the detection accuracy.
- The proposed incremental learning strategy improves the generalization capability of the proposed model. The comprehensive experiments show that our model can achieve good performance on a new dataset, while maintaining their performance on previous dataset.

## 2 RELATED WORK
### 2.1 Deepfake Detection

Several studies have been proposed in the past to detect forged media. Most of the proposed methods employ Convolutional Neural Networks (CNN) based approaches to detect deepfake video. However, the proposed techniques seem to struggle against newer deepfake detection benchmarks.

In [46], Rossler *et al.*, propose a diverse and high quality deepfake dataset, which they call, FaceForensicss++ dataset. They employ a simple Xception [12] network pre-trained on imagenet dataset, and fine-tune it on FaceForensics++ dataset. They report excellent performance scores on fake datasets (FaceSwap, Face2Face, DeepFakes, NeuralTextures), which are subsets of FaceForensicss++ dataset [46]. The detection models lack the generalization capabilities on real world data. In [30], Li *et al.*, propose novel image representation technique to detect forged face images [46].

Afchar *et al.* [1] proposed a face foregery detection network called MesoNet. They propose two networks Meso-4 and Meso Inception-4 with a small number of layers which focus on mesoscopic features in face images. They evaluate their networks on a public dataset [46] and a dataset they generate from videos available online. Ciftci *et al.* propose a deepfake video detection system and construct a deepfake dataset [13]. The proposed method employs biological signals hidden in portrait videos. The motivation is that the biological signals are neither temporally nor spatially conserved in manipulated videos. They extract remote photoplethysmography (rPPG) signals from various face parts and combine those features to train their models. It achieves better performance on deepfake video detection compared to image based detection methods. Since it relies on biological signals, which measures the subtle changes of color and motion in RGB videos, this approach has the potential to fail on facial images with different poses when evaluated on the portrait videos. Also the rPPG technique can be fooled by intentionally changing the skin tone in the post-processing stage of deepfake video generation.

In [27], Guera *et al.* proposed a pipeline which employs a CNN along with a long short term memory (LSTM) network to detect manipulated videos. The CNN backbone is used to extract frame-level features. The manipulated videos possess temporal inconsistencies among video frames that can be detected in a recurrent network. Sabir *et al.* [47] propose a recurrent convolutional networks to detect manipulated media with different backbones, including ResNet50, DenseNet and bidirectional recurrent network. The DenseNet backbone with face alignment and bidirectional recurrent network achieves the best performance.

Nguyen *et al.* [40] employ a model based on capsule networks to detect manipulated video. The proposed pipeline consists of pre-processing phase, a VGG-19 CNN backbone, capsule network and post ptocessing phase. Nguyen *et al.* [39] propose a different strategy to detect deepfake video. They use a multi-task convolutional neural network to detect and locate manipulated facial regions in videos and images. The proposed network comprises of an encoder and a Y-shaped decoder network. The encoder is used for binary classification. By fine-tuning the model on a small amount of data, it can deal with in-the-wild manipulated videos. Mittal *et al.* propose a multi-modal deepfake detection method in [38]. They use audio and visual information to train their models. The coherence between audio and visual modalities can be learned by the model. Additionally, the emotions extracted from the facial images are considered when detecting the deepfake videos. They train the model with triplet loss. Agarwal *et al.* [2] propose a deepfake detection system based on behavioral and appearance features. The behavioral embeddings can be learned using a CNN model by employing a metric-learning loss function. The model is tested on a number of different datasets including FaceForensicss++ [46], DeepFake Detection, DFDC [21], CelebDF [32] etc. The technique works for the face swapped deepfake videos, but they have the potential to fail in the detection of deepfake video that is generated using facial re-enactment and facial attribute manipulation techniques. The existing works focus on CNNs to detect deepfake video. A limited number of works [27, 47] use the recurrent networks which can process a video as a whole rather than image by image in deepfake video detection.

### 2.2 Transformer Architecture

The basic building block of transformer is the multi-head self-attention mechanism [50]. The self-attention mechanism is responsible for learning the relationship among the elements of input sequence. Transformer architectures can accommodate the full-length input sequences in a parallel manner and learn the dependency among frames. The transformer models can also be scaled to extremely complex models on large-scale datasets.

In the natural language processing tasks e.g., text classification, machine translation, question answering, transformers have achieved state-of-the-art performance, including BERT [20], Ro-BERT [33] and GPTv1-3 [7, 43, 44]. BERT-large model which had 340 million parameters was beaten by a considerable margin by the GPT-3 [7] model which had 175 billion parameters. At present, the state-of-the-art Switch transformer [23] can scale up to a gigantic 1.6 trillion parameters. Inspired by the success of Transformers in NLP tasks, we can employ these models for vision and multi-modal vision-language tasks.

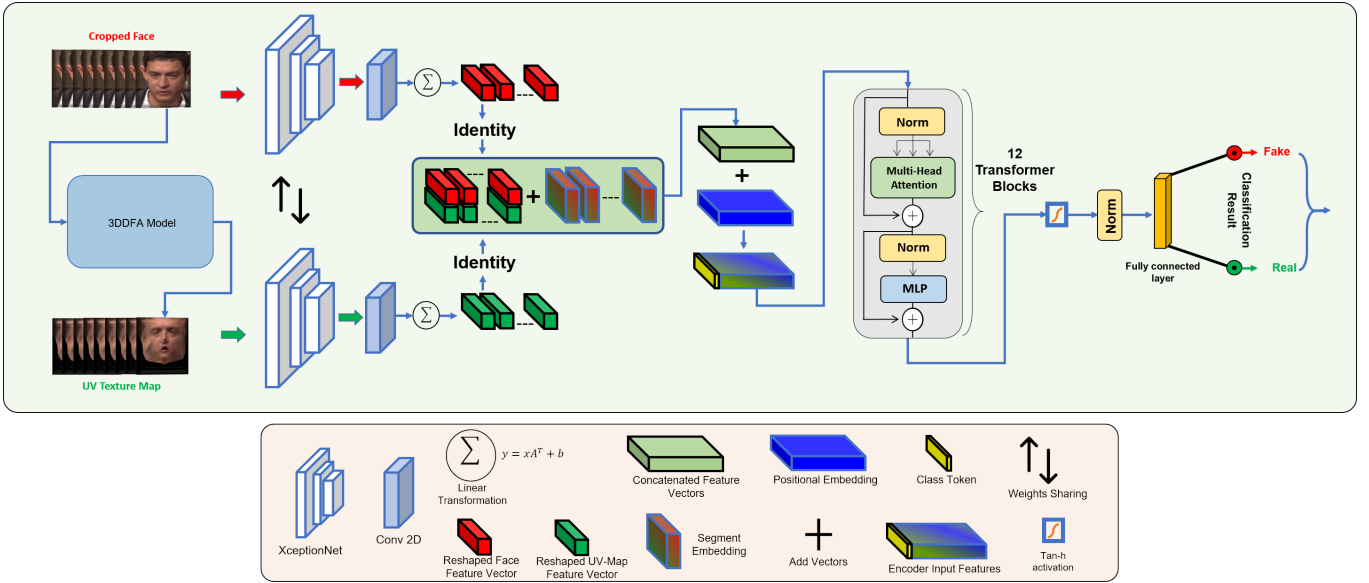

Figure 1: The architecture of the proposed video transformer, including the cropped face images and their corresponding UV texture maps as input, XceptionNet as backbone for image feature extraction and 12 transformer blocks for feature learning.

A large number of transformer based models are used to deal with the vision tasks, such as image classification[22], object detection [8], image segmentation [49], image captioning [34], video classification [24, 48], and visual question answering [11, 34, 42]. The transformer based models achieve stat-of-the-art performance in the vision tasks. The self-attention operation of the transformer architecture scales quadratically, which becomes enormously expensive as the length of the input sequence increases. A number of more efficient transformer architectures are used to address this issue [4, 15, 35, 49, 52]. Specifically, the transformer based methods achieve state-of-the-art performance in image classification tasks.

## 3 METHODOLOGY

In the following sections, we introduce the backbone to extract image features, the proposed video transformer model and the proposed incremental learning strategy.

### 3.1 Backbone

Inspired by Vision Transformer [22], the high level image features are more informative than the image patches. Thus, we employ a pre-trained CNN backbone to extract image features. Since XceptionNet achieves better performance than other backbone networks in deepfake detection, we employ XceptionNet [12, 53] as the image feature extractor.

We employ a Single Stage Detector (SSD) to detect and crop faces frame by frame [6]. We use 3D Dense Face Alignment (3DDFA) model [25, 26] to generate UV texture maps from face images. We use both face images and their UV texture maps to extract the image features. The existing methods use aligned 2D face images. Such an alignment only centralizes the face without considering whether the face is frontalized. When the face is not frontalized, the face part that is not captured by the camera can cause facial information loss and misalignment with the face images that are frontalized. With the UV texture, all face images are aligned into the UV map that is created from the generated 3D faces. Since the generated 3D faces cover all the facial parts, there is no information loss. In UV map, the facial part for all the faces can be located in the same spatial space. The aligned face image can also provide pose, eyes blink and mouth movement information that cannot be perceived in the UV texture image, so we use both face images and their UV texture maps to extract the image features.

### 3.2 Video Transformer

Figure 1 illustrates the architecture of the proposed video transformer for deepfake detection. To learn the intra-frame dependencies, we train our model a sequence of the cropped facial images with their UV texture maps, as illustrated in Figure 1. We employ pre-trained XceptionNet to extract feature maps from face images and the UV texture maps. After getting the feature maps of each face image frame and the corresponding UV texture map, we reshape the feature vectors using a 2D convolution layer and a linear layer to accommodate the input dimension of video transformer.

In the proposed video transformer model, we exploit the property of parallel input processing, which is inherent in the transformer models. We feed a sequence of facial images and their corresponding UV texture maps to the video transformer model. The recurrent networks can learn to detect the intra-frame discrepancies, such as flickering, blurry frames, and mouth movement [27, 47]. The cropped face image and its corresponding UV texture map are used as the input to the XceptionNet backbone. The single face frame and UV texture map can be represented as:

$$\mathbf{f} \in \mathbb{R}^{(\frac{N}{2T}) \times D} \quad (1)$$

$$\mathbf{u} \in \mathbb{R}^{(\frac{N}{2T}) \times D} \quad (2)$$

where $f$ represents face feature vector, and $u$ represents UV texture map feature vector, $N$ represents the total number of patches. For facial image and UV texture map, $N$ is 576, while $N$ is 324 for facial image only. $T$ represents the number of input frames and $D$ represents the constant latent vector dimension. We concatenate each face image frame and the corresponding UV texture map as a feature vector:

$$(\mathbf{f,u}) \in \mathbb{R}^{(\frac{N}{T}) \times D} \quad (3)$$

We use one dimensional learnable segment embeddings to help our model distinguish different types of inputs in the input data structure. We add segment embeddings to the feature vector which results from the fusion of facial images and their corresponding UV texture map feature vectors as shown in Equation 3. The segment embeddings can be defined as:

$$\mathbf{E}_{seg} \in \mathbb{R}^{(\frac{N}{T}) \times D} \quad (4)$$

The input feature vector to transformer can be extracted from the concatenation of the facial frame and its corresponding UV map:

$$frame_i = [((face_0...face_{\frac{N}{2T}}), (uv_0...uv_{\frac{N}{2T}})) \times D] \\ + \mathbf{E}_{seg-face-uv_{\frac{N}{T}}} \quad (5)$$

where $N = 576$, $T = 9$ and $D = 768$.

A learnable positional embedding is then added to the whole input feature vector:

$$\mathbf{E}_{pos} \in \mathbb{R}^{(N+1) \times D} \quad (6)$$

The video transformer model consists of 12 encoders, where each encoder includes a multi-head self-attention layer, two Norm layers and a Multi-Layer Perceptron (MLP).

### 3.3 Incremental Learning

We use an incremental learning strategy to fine-tune the proposed model on new datasets, without sacrificing its performance on previous datasets. The loss function in incremental learning consists of two parts: one part that measures the similarity between the weights from a new dataset and the old weights from the previous dataset, and the other one is to measure the accuracy of the training model on the new dataset [9]. The former one forces the weights to be as similar as possible to the old weights, so it still performs well on the previous dataset. And the latter one guarantees that the model performs well on the new dataset. We first train the proposed model on FaceSwap and Deepfakes subsets in FaceForensics++ dataset [46] that are generated using faceswap technique. Then we fine-tune the model on the other two subsets of the FaceForensics++ dataset, Face2Face and Neural Textures, which are generated by a different technique called facial re-enactment [46]. To show the performance on unseen dataset, we also fine-tune the model on DFDC dataset [21] and DeepFake Detection (DFD) dataset [46] as shown in Figure 2. We train the model with segment embeddings on 280k images from FaceSwap and Deepfakes subsets of FaceForensics++ dataset. The trained model is fine-tuned on only 2,500 images from the Face2Face subset which are 0.05% of the Face2Face subset. The trained model is finetuned on 2,500 images from Neural Textures subset. Then we can fine-tune the trained model from the previous step on 2,500 images from DFD dataset [46]. Finally, we fine-tune the trained model on 6,000 DFDC images, which are also 0.05% of the DFDC dataset [21].

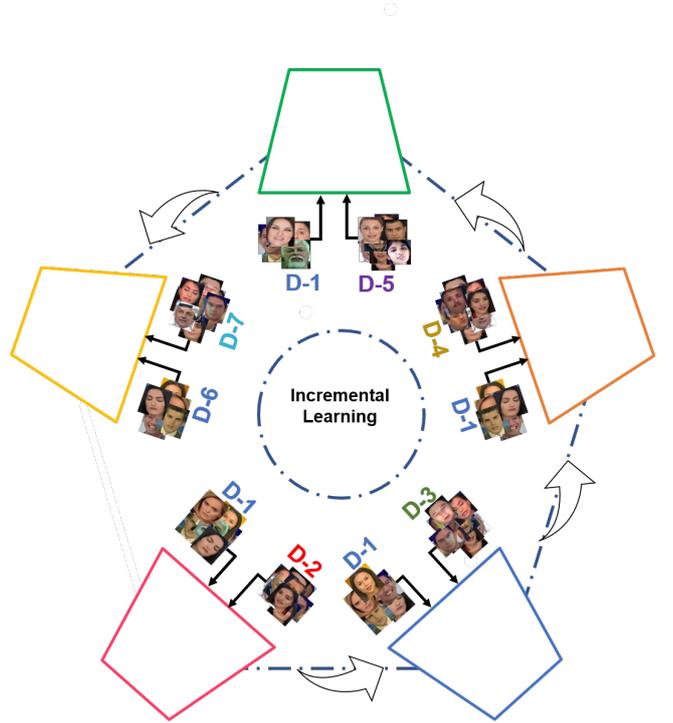

Figure 2: Illustration of the proposed incremental learning strategy. D1 represents the real data used to train the models. Whereas, D2 comprises of FaceSwap and Deepfakes datasets. D3 represents the Face2Face dataset and D4 represents Neural Textures dataset. D5 and D6 represents DFDC dataset and D7 represents DeepFake Detection (DFD) dataset.

| Dataset | Training | Validation | Test |
|---|---|---|---|
| Prestine | 138000 | 27600 | 1400 |
| FaceSwap | 69000 | 13800 | 1400 |
| Deepfakes | 69000 | 13800 | 1400 |
| Face2Face | 2500 | 500 | 1400 |
| Neural Textures | 2500 | 500 | 1400 |
| DFDC | 6000 | 1200 | 3500 |
| DFD | 2500 | 500 | 1400 |

Table 1: Number of frames used to train models on different datasets: Prestine, FaceSwap, Deepfakes, Face2Face, Neural Textures, DFDC and DFD.

## 4 EXPERIMENTS

### 4.1 Datasets

We train and evaluate our models on public deepfake detection benchmark, FaceForensics++ [46]. The FaceForensics++ dataset includes four different subsets: (1) FaceSwap, (2) Deepfakes, (3) Face2Face and (4) Neural Textures. The first two subsets contain videos generated by the face swapping techniques, whereas the other two subsets are generated by the facial re-enactment techniques. There are 1,000 videos in each subset. The FaceForensics++ benchmark also contains 1,000 real videos. We use 720 videos from each subset for training and 140 videos for validation and 140 videos for testing. The FaceForensics++ dataset contains around 1.7 million frames. We use FaceSwap and Deepfakes subsets for model training by employing only 280k frames for training. We further fine-tune these models on Face2Face and Neural Textures subsets. The trained models are also finetuned on DFDC dataset [21] and DFD dataset [46]. Table 1 shows the exact number of frames used to train and fine-tune the models from each dataset.

### 4.2 Implementation Details

For face detection, we employ Single Shot Detector (SSD) with ResNet as backbone. We employ 3DDFA-V2 [25, 26] to generate UV texture maps [18]. We use XceptionNet [12] for image feature extraction. We employ transformer architecture [22], including 12 transformer layers. We modify the Vision Transformer (ViT) base architecture [36] by adding the learnable segment embeddings to the input data structure. It enables the learning of the visual details in image frames with temporal information.

The raw input size of the images we feed to our model is [3, 299, 299]. We use this size to make our inputs compatible to the backbone network XceptionNet that is used for image feature extraction in our hybrid models. After we extract image features through XceptionNet, we get a feature vector of dimension [2048, 10, 10], we then pass this feature vector to a 2D convolutional layer and a linear layer, which gives us a feature vector of dimension [1, 32, 768]. After we get the two reshaped feature vectors for facial images and UV texture maps, we concatenate these two feature vectors and get a feature vector of dimension [1, 64, 768]. We add one dimensional learnable segment embedding to the feature vector as [1, 64, 768]. We do this for all the sequence frames and concatenate them as [1, 576, 768]. We add the learnable positional embeddings to the feature vector and a [class] token is added at the beginning of the feature vector. We use the final feature vector [1, 577, 768] as the input to the proposed video transformer model. We train all models for 5 epochs, with a learning rate of $3 \times 10^{-3}$. We choose SGD as optimizer, and use CrossEntropyLoss as the loss function.

### 4.3 Ablation Study

We present an ablation study with different experimental settings to show the effectiveness of the proposed modules. We train the models with 8 different configurations:

(1) Patch embedding transformer trained on face images only
(2) Patch embedding transformer trained on face and UV textures without segment embeddings
(3) Patch embedding transformer trained on face and UV textures with segment embeddings
(4) Hybrid image transformer trained on face images only
(5) Hybrid transformer trained on face and UV textures without segment embeddings
(6) Hybrid transformer trained on face and UV textures with segment embeddings
(7) Hybrid video transformer trained on 9 frames (face + UV texture maps) without segment embeddings
(8) Hybrid video transformer trained on 9 frames (face + UV texture maps) with segment embeddings

The listed models are trained and evaluated on 2 subsets of FaceForensics++ dataset: FaceSwap and Deepfakes. We train our models on around 280k images. The performance comparison is shown in Table 2.

*4.3.1 Patch embedding transformer trained on face images only.* The patch embedding based models are trained on 2D image patches. The first model is trained on face images only, and no UV texture maps. The input image is reshaped into 2D patches as $\mathbf{x}_{input} = \mathbf{f} + \mathbf{E}_{pos}$, $f$ represents the reshaped face frame image. $N$ refers to the number of patches, which is *324* in our case. $D$ represents the constant latent vector dimension, which is *768* in our model. After adding a BERT styled [*class*] token at the beginning of our input, the dimension of the input feature vector is [1, 325, 768].

*4.3.2 Patch embedding transformer trained on face and UV textures without segment embeddings.* This model is trained on 2D patches of facial images and the UV texutre maps without adding the learnable segment embeddings to the input data structure. We only use the positional embeddings, so we can compare this model to the model with both the positional embeddings and the segment embeddings. As illustrated in the first model and the second model of Table 2, it shows that the UV texture map provides useful information for deepfake detection in the patch embedding transformer models.

*4.3.3 Patch embedding transformer trained on face and UV textures with segment embeddings.* We add one dimensional learnable segment embeddings with positional embeddings to train this model. The purpose of adding segment embeddings is to help model distinguish the face image patches and the UV texture map patches. As shown in the second model and the third model of Table 2, We can see that the model trained with the segment embeddings performs better than the model trained without segment embeddings. This implies that the proposed segment embeddings help enhance the feature learning, thereby improving the detection performance.

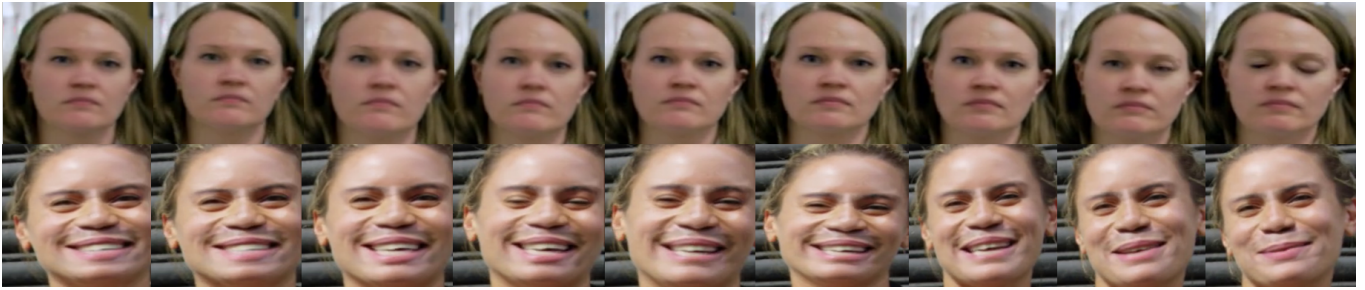

Figure 3: Frames of two example deepfake videos from DeepFake Detection (DFD) dataset. The image transformer model classifies the videos as "Real", while the video transformer correctly classifies both videos as "Fake".

| Models | FaceForensics++ | | |
|---|---|---|---|
| | AUC | F1-Score | Accuracy |
| Patch Img Only | 74.39% | 65.88% | 72.31% |
| Patch UV Img | 66.48% | 56.79% | 66.58% |
| Patch SE UV Img | 77.10% | 68.71% | 73.26% |
| Hyb Img Only | 98.03% | 96.19% | 97.37% |
| Hyb UV Img | 98.57% | 97.20% | 98.09% |
| Hyb SE Img | 99.28% | 98.92% | 99.28% |
| Video | 98.74% | 97.54% | 98.32% |
| Video SE | **99.64%** | **99.28%** | **99.52%** |

Table 2: Detection accuracy of different models on FaceForensics++ dataset (FaceSwap and Deepfakes). Patch Img Only, Patch UV Img, Patch SE UV Img, Hyb Img Only, Hyb UV Img, Hyb SE Img, Video, Video SE are corresponding to the models described in Section 4.3.1 to 4.3.8, respectively.

*4.3.4 Hybrid image transformer trained on face images only.* The hybrid image transformer model are trained on image features extracted from face image only using the XceptionNet backbone. As can be seen from Table 2, the hybrid model outperforms the patch embedding based model. So the image feature backbone is necessary in the transformer based deepfake detection model.

*4.3.5 Hybrid transformer trained on face and UV textures without segment embeddings.* Hybrid transformer model for face images and UV texture maps is trained without the learnable segment embeddings. The results in Table 2 show that the UV texture map provides useful information for deepfake detection in the hybrid transformer models. This lies in the fact that the UV texture map is losslessly better aligned than the aligned face image. The aligned face image also provides pose, eyes blink and mouth movement information that cannot be perceived in the UV texture image.

*4.3.6 Hybrid transformer trained on face and UV textures with segment embeddings.* This hybrid image transformer model is trained using segment embeddings. When we compare the hybrid image transformer model with and without the segment embeddings as shown in Table 2, We can see that the model trained with the segment embeddings performs better than the model trained without the segment embeddings. So the segment embeddings help the model distinguish the two different types of the input data, thereby enhancing the feature learning in the transformer.

*4.3.7 Hybrid video transformer trained on facial image frames (face + UV texture maps) without segment embeddings.* The structure of the video based transformer model is different from the image only based transformer models as described above. The video based models are fed with the consecutive face images and their corresponding UV texture maps. We train this video transformer model on the face image frames and their corresponding UV texture maps without adding the proposed segment embeddings. We only add the positional embeddings to the input sequence of the face frames and their corresponding UV maps. The video based transformer performs better compared to the image only based transformer model without the segment embeddings.

*4.3.8 Hybrid video transformer trained on facial image frames (face + UV texture maps) with segment embeddings.* The hybrid video transformer achieves the best performance among all the experimental settings. We train this model on the consecutive face image frames along with their corresponding UV texture maps. We add both the segment embeddings and the positional embeddings to the input data structure. We add separate embeddings to each of the input face frames and each of the corresponding UV texture maps. It helps the video transformer model to discriminate the input frames and achieve better performance as shown in Table 2. Figure 3 shows frames of two example deepfake videos from DeepFake Detection (DFD) dataset. The image transformer model classifies the videos as "Real", while the video transformer correctly classifies both videos as "Fake".

## 4.4 Incremental Learning

The performance of models with incremental learning is shown in Table 3. We first train the models on 280k images from FaceSwap and Deefakes subsets from the FaceForensics++ dataset. We finetune the models on four different datasets: (1) Face2Face [46], (2) Neural Textures [46], (3) DFD [46] and (4) DFDC [21].

We use less than 0.5% of original data to finetune the models with incremental learning. More specifically, we use 2500 images to finetune the model on Face2Face, 2500 images on Neural Textures, 2500 images on DFD dataset, and 6000 images on DFDC dataset. Note that the DFDC dataset includes around 1.5 million frames. Table 3 shows that the proposed models finetuned on a small amount of data can still achieve good performance on new datasets, while maintaining their performance on the previous datasets. The main reason lies in the loss function in incremental learning. It consists of two parts: one part that measures the similarity between the weights from a new dataset and the old weights from the previous dataset, and the other one is to measure the accuracy of the training model on the new dataset. The former one forces the weights to be as similar as possible to the old weights, so it still performs well on the previous dataset. And the latter one guarantees that the model performs well on the new dataset.

| Fine-Tuning Dataset | FS | DF | Pristine | F2F | NT | DFD | DFDC | Commultive Accuracy |
|---|---|---|---|---|---|---|---|---|
| F2F | 95.00% | 99.28% | 100.00% | 99.28% | - | - | - | 98.39% |
| Neural Textures | 96.42% | 99.28% | 100.00% | 100.00% | 94.28% | - | - | 98.00% |
| DFD | 97.85% | 100.00% | 98.56% | 98.57% | 90.00% | 99.28% | - | 97.38% |
| DFDC | 93.57% | 92.14% | 76.97% | 88.57% | 51.42% | 93.27% | 91.69% | 83.95% |

Table 3: Incremental learning strategy. Performance of hybrid image transformer trained with segment embeddings on FaceSwap and Deepfakes datasets and finetuned on Face2Face, NeuralTextures, DFD dataset and DFDC dataset.

| Method | Dataset | | |
|---|---|---|---|
| | FF++ | DFD | DFDC |
| Rossler et.al [46] | 95.73% | 88.07% | 85.60% |
| Mittal et.al [38] | - | - | 84.40% |
| Zhu et.al [56] | 99.61% | 89.84% | 87.93% |
| Li et.al [30] | - | 93.34% | 73.52% |
| Bonettini et.al [5] | - | 89.35% | 85.71% |
| Guera et.al [27] | 83.10% | - | - |
| Ours (Image+Video Fusion) | **99.79%** | **99.28%** | **91.69%** |

Table 4: Performance comparison with other deepfake detection methods on FaceForensics++ (FF++) dataset, DFD dataset and DFDC dataset.

| Method | Accuracy | Num. Train Images |
|---|---|---|
| Rossler et.al [46] | 98.36% | 870k |
| Afchar et.al [1] | 84.56% | 9k |
| Zhu et.al [56] | 98.22% | 172k |
| Li et.al [30] | 98.64% | - |
| Ours | **99.28%** | **5k** |

Table 5: Performance comparison on Face2Face dataset. We finetune the models on the different number of images from Face2Face dataset.

## 4.5 Comparison

We compare the results achieved by the proposed models with state-of-the-art deepfake detection systems. Image Transformer refers to the model trained with the settings as described in Section 4.3.6 and Video Transformer refers to the model trained with the settings as described in Section 4.3.8. In Table 4, we demonstrate the results of fusing the predictions from Image Transformer and Video Transformer by averaging the probabilities from both models to get the final output score. The fused models outperform state-of-the-art deepfake detection systems on FaceForensics++ dataset, DFD dataset and DFDC datasets.

In Table 5, our model outperforms state-of-the-art detection systems when trained and tested on a specific subset of FaceForensics++ dataset: Face2Face. We can finetune the model on a smaller amount of data and achieve better performance when compared to other methods as shown in Table 5. This demonstrates a more enhanced generalization capability of the video transformer model with incremental learning.

## 5 CONCLUSION

In this paper, we propose a video transformer with incremental learning for deepfake detection. The novel design in video transformer enables the informative feature learning, thereby improving the performance in deepfake detection. The proposed incremental learning strategy enhances the generalization capability of the proposed model. Experimental results on various public datasets demonstrate that our method outperforms stat-of-the-art methods, and each component in our method is effective with significant performance gains.